\documentclass{INTERSPEECH2023}

\usepackage{amsmath}
\usepackage{amssymb}
\usepackage{graphicx}
\usepackage{hyperref}
\usepackage{float}
\usepackage{booktabs}  
\usepackage{xcolor}   
\usepackage{listings} 
\usepackage{caption}  
\usepackage{multirow}
\usepackage[utf8]{inputenc}
\usepackage{cuted}

\definecolor{backcolour}{HTML}{F5F5EF}
\definecolor{commentcolour}{HTML}{008000}
\definecolor{keywordcolour}{HTML}{D7196F}
\definecolor{stringcolour}{HTML}{008000}

\renewcommand{\thefootnote}{\fnsymbol{footnote}}

\DeclareCaptionFont{captionfont}{\fontsize{8.pt}{10pt}\selectfont}
\interspeechcameraready

\title{Context-Aware Interleaved Batching for WhisperX} 
\name{Carlos Bain$^*$, Max Bain$^\dagger$} 
\address{
  University of Oxford$^\dagger$}

\makeatletter
\renewcommand{\@maketitle}{%
  \newpage
  \begin{center}%
    {\Large \bf \@title \par}%
    \vskip 1.5em%
    {\large
     \begin{tabular}[t]{c}%
       \@name \\
       \@address
     \end{tabular}\par}%
    \vskip 1em
    \includegraphics[width=0.99\textwidth,height=5.2cm,keepaspectratio]{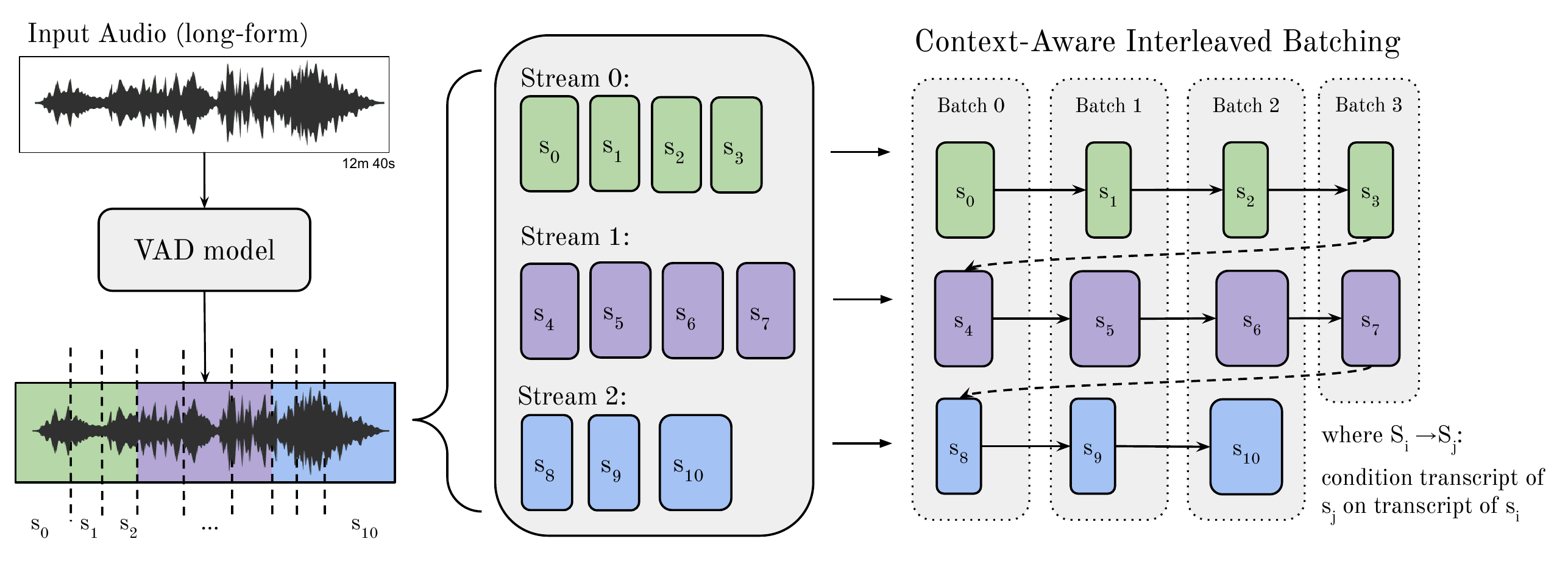}\par
    \vspace{2mm}
    \begin{minipage}{0.98\textwidth}
      \captionof{figure}{\textbf{Context-Aware Interleaved Batching for speech transcription.} Long-form audio is first segmented into chunks ($\le$30s) via Voice Activity Detection (VAD). These chunks are partitioned into contiguous streams and batched by drawing one chunk per stream. To preserve text context during batched transcription, each chunk is conditioned on a FIFO rolling text buffer generated from its stream's preceding chunks. Finally, to resolve the cold-start problem, the first batch is re-transcribed using the final text of chronologically preceding streams as context.}
      \label{fig:redo_logic}
    \end{minipage}%
  \end{center}%
  \par
  \vskip 1.4em
}
\makeatother

\begin{document}

\maketitle
\renewcommand{\thefootnote}{\fnsymbol{footnote}}

\footnotetext[1]{Correspondence at \texttt{carlos.o.bain@gmail.com}}
\footnotetext[2]{Work done whilst at University of Oxford, now currently at Google DeepMind.}

\begin{abstract}
While WhisperX accelerates speech transcription via intra-audio batching, it isolates audio segments, losing the historical context needed for coherent punctuation and terminology transcription. Conversely, standard Whisper retains context sequentially but suffers from slow inference and hallucination loops. To achieve the best of both worlds, we propose Context-Aware Interleaved Batching. By using VAD-derived segment boundaries, our algorithm stabilizes Whisper's text conditioning, allowing us to safely maintain continuous historical context across batched audio segments. As demonstrated on long-form audio benchmarks, this approach reduces Word Error Rate (WER) and improves proper noun transcription, all while maintaining high-throughput inference speeds.
\end{abstract}

\section{Introduction}

Large-scale speech recognition models, such as OpenAI’s Whisper \cite{radford2022}, achieve high accuracy across diverse domains. However, deploying these models on long-form audio is computationally expensive. Standard Whisper processes audio sequentially, preventing batched inference. Furthermore, while Whisper includes a \texttt{condition\_on\_previous\_text} feature to maintain historical context, it is often unstable. As demonstrated in our evaluations, enabling this feature in standard Whisper can actually degrade overall Word Error Rate (WER), a known symptom of the model hallucinating or propagating errors across sequential forward passes~\cite{peng2023reproducing}. In addition, recent work~\cite{ahn2026} further highlights the risks of autoregressive context in long-form transcription, showing that an erroneous context can propagate across successive decoding steps and trigger hallucinations. 

To resolve this, WhisperX~\cite{bain2023} introduced a Voice Activity Detection (VAD) based pipeline. By segmenting audio at natural silences rather than arbitrary fixed-length windows, WhisperX offers significant inference speedups via within-audio batched transcription, improves overall transcription accuracy, and adds highly accurate word-level timestamps via forced alignment with a CTC model~\cite{kurzinger2020ctc}. Despite these advantages, batched processing inherently isolates each audio chunk. This prevents the model from conditioning on previous text, stripping it of its autoregressive context capabilities. On long-form audio, this loss of historical context degrades performance on continuous conversational threads, technical language, and proper nouns.  Prior works in end-to-end ASR have long established that such context, via contextual biasing, is essential for accurately transcribing rare and domain-specific terms~\cite{williams2018contextual,Jain2020ContextualRF,Chang2021ContextAwareTT}.

Currently, no transcription method simultaneously achieves high parallelization, reliable word-level timestamps, and stable autoregressive conditioning. We address this with Context-Aware Interleaved Batching. By propagating historical text across parallel batch boundaries, our method maintains continuous context without sacrificing inference speed. Crucially, decoupling audio chunk boundaries from Whisper's internal timestamp decoding prevents hallucination loops, stabilizing the model's conditioning on previous text and improving accuracy.

Our contributions are threefold: (i) a novel method enabling historical text conditioning within fully parallelized transcription frameworks; (ii) a new, open-source medical long-form benchmark, evaluated alongside targeted Proper Noun (Pn) and punctuation metrics; and (iii) empirical demonstration that our method improves accuracy across all metrics and integrates into any batched, segmentation-based transcription pipeline.

\section{Method}

In this section, we describe the parallel speech transcription baseline introduced by WhisperX~\cite{bain2023}, followed by our proposed Context-Aware Interleaved Batching method.

\subsection{Overview of Baseline Architectures}

We first review the autoregressive conditioning mechanism of the standard Whisper model~\cite{radford2022}. Whisper operates as an encoder-decoder Transformer. During standard sequential inference, the audio is processed in fixed windows up to 30 seconds. When transcribing a given window $i$, the model relies on two distinct attention mechanisms to generate text:

\begin{enumerate}
    \item \textbf{Cross-Attention:} The encoder processes the audio features of window $i$ and passes the representations to the decoder.
    \item \textbf{Self-Attention (Conditioning):} Because Whisper processes these audio windows chronologically, the decoded text tokens from the previous window $i-1$ can be provided as a conditioning prefix to the decoder for window $i$.
\end{enumerate}

To determine the boundaries of the input audio chunks for subsequent passes, standard Whisper autoregressively decodes specialized timestamp tokens. By predicting the end time of the current speech segment, the model dictates how far to shift the 30-second audio window for the next forward pass. However, this internal timestamp decoding is prone to errors and hallucination, frequently causing the model to predict incorrect audio boundaries, which leads to erroneous window shifting, looping, and severe text hallucinations. Furthermore, the \texttt{condition\_on\_previous\_text} feature acts as a First-In, First-Out (FIFO) rolling buffer, retaining up to 224 previously decoded text tokens as a historical prefix for the next chunk. While designed to maintain conversational continuity, feeding a 224-token context derived from erroneous timestamp boundaries often traps the model in a negative feedback loop, propagating hallucinations forward.\\

WhisperX circumvents this fragile timestamp decoding by relying on a robust Voice Activity Detection (VAD) pipeline. The long-form audio is first segmented by a lightweight VAD model~\cite{bredin2020pyannote} at natural silences, rather than relying on the Whisper model's internal text-based timestamps. This results in a temporally ordered sequence of $N$ variable-length chunks, denoted as $\mathbf{s} = [s_0, s_1, \dots, s_{N-1}]$. These individual chunks (each up to 30 seconds) are then grouped into batches and transcribed in parallel, enabling significant inference speedups, as well as improved word-level timestamps and overall transcription accuracy. However, batched processing isolates each chunk. If chunks $s_{i-1}$ and $s_i$ are processed simultaneously in the same batch, the text tokens of $s_{i-1}$ cannot be passed as a prefix to $s_i$. Without this text conditioning, the model loses contextual awareness across chunk boundaries. This frequently results in textual inconsistencies between segments, particularly regarding punctuation, capitalization, and the transcription of proper nouns.

\subsection{Context-Aware Interleaved Batching}

To retain both the autoregressive text-conditioning of standard Whisper and the throughput of parallel transcription, we propose Context-Aware Interleaved Batching. Given the VAD-segmented sequence $\mathbf{s}$ and a target hardware batch size $B$, we partition the sequence into $B$ contiguous streams of length $M = \lceil N / B \rceil$ (padding the final stream with empty audio if $N$ is not divisible by $B$). The $b$-th stream, for $b \in \{0, 1, \dots, B-1\}$, is defined as:

\begin{equation}
    \mathbf{S}_b = [s_{b \cdot M},\, s_{b \cdot M + 1},\, \dots,\, s_{b \cdot M + M - 1}]
\end{equation}

Transcription proceeds over $M$ sequential batch iterations. At step $m \in \{0, 1, \dots, M-1\}$, we construct batch $\mathcal{B}_m$ by extracting the $m$-th chunk from every stream:

\begin{equation}
    \mathcal{B}_m = \left\{ \mathbf{S}_b[m] \mid b \in \{0, 1, \dots, B-1\} \right\}
\end{equation}

Because chunk $\mathbf{S}_b[m]$ immediately follows $\mathbf{S}_b[m-1]$ in the original audio, the decoded text from batch $\mathcal{B}_{m-1}$ is appended to a stream-specific FIFO rolling context buffer (capped at 224 tokens). This rolling buffer then serves as the conditioning prefix for the corresponding chunk in batch $\mathcal{B}_m$. This enables standard Whisper's autoregressive context passing across the audio while safely operating on isolated VAD boundaries, all while processing $B$ chunks in parallel.\\

\textbf{Cold-Start Resolution.}
At the start of each stream ($m=0$), the preceding chronological chunk is $\mathbf{S}_{b-1}[M-1]$, which belongs to the final batch $\mathcal{B}_{M-1}$. Since $\mathcal{B}_{M-1}$ is not yet transcribed when evaluating $\mathcal{B}_0$, the initial batch $\mathcal{B}_0$ is processed without text conditioning. Once all $M$ batches are transcribed, $\mathcal{B}_0$ is re-transcribed using the newly generated text from $\mathcal{B}_{M-1}$ as the context prefix, ensuring consistent context boundaries across all streams.

\definecolor{deepblue}{rgb}{0,0,0.5}
\definecolor{deepred}{rgb}{0.6,0,0}
\definecolor{deepgreen}{rgb}{0,0.5,0}
\definecolor{backcolour}{rgb}{0.98,0.98,0.96}

\lstdefinestyle{pythonstyle}{
    language=Python,
    backgroundcolor=\color{backcolour},   
    commentstyle=\color{deepgreen}\itshape,
    keywordstyle=\color{deepblue}\bfseries,
    stringstyle=\color{deepred},
    basicstyle=\ttfamily\scriptsize, 
    breakatwhitespace=false,         
    breaklines=true,                 
    captionpos=b, 
    keepspaces=true,                 
    numbers=left, 
    numbersep=5pt,                  
    numberstyle=\tiny\color{gray},
    showspaces=false,                
    showstringspaces=false,
    showtabs=false,                  
    tabsize=4
}
\begin{lstlisting}[style=pythonstyle, caption={Pseudocode for Context-Aware Interleaved Batching for WhisperX transcription.}, label={lst:batched_transcription}]
def batched_transcription(audio, batch_size, max_ctx=224):
    '''
    audio - input audio signal
    batch_size - number of streams (B)
    max_ctx - maximum number of text tokens for conditioning
    returns:
        segs - chronological array of transcript segments
    '''
    # Extract speech chunks <= 30s
    chunks = extract_vad_chunks(audio)
    
    # Split into B sequential streams
    streams = distribute_contiguous(chunks, batch_size)
    
    # Group across streams for batching
    batches = interleave_roundrobin(streams)
    
    # Initialise empty rolling ctx per stream
    ctx = [[] for _ in range(batch_size)]
    segs = []

    for b in batches:
        out = batch_transcribe(b, ctx)
        segs.append(out)
        
        # update rolling context
        for sdx in range(len(out)):
            # Extract token IDs to feed into
            # the next forward pass
            ctx[sdx] += out[sdx]['tokens']
            # Truncate to the max ctx window
            ctx[sdx] = ctx[sdx][-max_ctx:]
            
    # Cold start resolution: stream i gets stream i-1th ctx.
    ctx_wrap = [None] + ctx[:-1]
    segs[0] = batch_transcribe(batches[0], ctx_wrap)

    # Flatten batches and sort chronologically 
    return sort_by_time(segs)
\end{lstlisting}

\section{Results}

\subsection{Evaluation Methodology}

\subsubsection{Evaluation Datasets}

\noindent\textbf{Earnings-21~\cite{delrio2021}.}
To evaluate our method, we use two benchmarks.
First, the widespread Earnings-21 corpus, a challenging dataset consisting of long-form, unscripted financial earnings calls from the year 2020. This corpus is particularly well-suited for benchmarking real-world transcription systems due to its high density of complex financial terminology, proper nouns, and continuous speech segments. Crucially, the Earnings-21 dataset provides strict ground truth punctuation markers, enabling us to evaluate punctuation accuracy.\\

\noindent\textbf{MedicalLessons.}
Second, we introduce a custom corpus of medical terminology lessons: \textit{}{MedicalLessons} - a purpose-built dataset consisting of educational lecture recordings covering foundational and anatomical medical vocabulary. The corpus contains a structured series of lessons covering topics such as medical prefixes and suffixes, anatomical terminology, sourced from YouTube. Ground-truth transcripts for all six recordings were manually verified to ensure correctness. This is released as an open source dataset\footnote{\url{https://huggingface.co/datasets/litosway/MedicalLessons}}.

\subsubsection{Metrics}

\noindent\textbf{WER.} To measure transcription accuracy we use the standard word error rate metric (WER).

\noindent\textbf{F1.} For punctuation we use F1; punctuation alignment between reference and hypothesis transcriptions is computed using a Levenshtein distance-based algorithm to isolate and evaluate grammatical pauses and structural errors independently of the normalized text. All metrics are computed at the transcript level rather than as averages over individual segments. For WER, all predicted segments within a file are concatenated into a single hypothesis string and all ground-truth segments are concatenated into a single reference string; WER is then computed once on these full-length strings using standard Levenshtein-based alignment.\\

\noindent Both of these metrics of course have limitations in their exact matching nature, and lack of ``semantic'' measurement. Standard ASR metrics have not been shown to measure proper nouns, technical language and punctuation properly~\cite{parulekar2025laserllmbasedasrscoring}. Therefore we additionally introduce LLM-based metrics.\\

\noindent\textbf{LLM Punctuation Score.} To calculate punctuation accuracy based on semantic meaning rather than rigid exact matches, we introduce the LLM Punctuation Score (LPS), similar to the LLM as judge metric shown in \cite{parulekar2025laserllmbasedasrscoring}. LPS ensures that stylistic choices, such as using a dash instead of a comma or a period instead of a clause-separating comma are not penalised. To compute the LPS, ground truth and predicted transcripts are stripped of casing and punctuation for word-level alignment using the jiwer library, the punctuation is separately tagged. These fully punctuated texts are divided into segments of exactly 120 GT words. Each paired segment is then passed to GPT-5.1 to evaluate the semantic accuracy of the predicted punctuation on a 1-to-3 scale. The prompt instructs the model to ignore word-level transcription errors and penalise only structural divergences that distort the core meaning. The final LPS for a predicted transcript is the mean of these individual segment scores. Given a transcript divided into $N$ segments, where each segment $i$ receives a score $S_i \in \{1, 2, 3\}$, the overall score is computed as $\text{LPS} = \frac{1}{N} \sum_{i=1}^{N} S_i$.\\ 

\noindent\textbf{Proper Noun Score (Pn).} For measuring adherence to proper nouns and domain-specific technical terms , we propose Proper Noun Score (Pn). Each ground-truth word is labeled as a proper noun or technical term by an LLM-based classifier. Word-level alignment between the ground-truth and
candidate transcript is then used to identify substitutions and deletions. The Pn Score measures the proportion of proper nouns that were correctly transcribed:
\begin{equation}
    \mathrm{Pn} = \frac{P - M}{P} \times 100
\end{equation}
\noindent where $P$ is the total number of proper nouns in the ground truth and $M$ is the number of those that were incorrectly transcribed (substituted or deleted). A score of 100 indicates perfect proper-noun accuracy. This metric is motivated by the limitations of conventional WER for capturing the significance of transcription errors. Semantic-WER~\cite{roy2021} addresses this limitation by assigning greater importance to semantically significant errors rather than treating all word errors equally. More recently, Bañeras-Roux et al.~\cite{banerasroux2026} investigate generative LLMs as ASR evaluators, showing that LLM-based evaluation can achieve substantially higher agreement with human judgments than WER when grading between competing transcription hypotheses. To better evaluate transcriptions, our Pn Score targets domain-specific technical terms and proper nouns, a superior way of measuring transcription quality.

\subsubsection{Implementation Details}
All evaluations were conducted using the  \texttt{openai/whisper-large-v2} as the base transcription model, executing on a single Nvidia T4 GPU. The default batch size of 16 was used on both WhisperX variants for the earnings data set. For the medical dataset, a batch size of 4 was used. Moving on, across both datasets compute type float16 was applied to all variants. Decoding used beam search (width 5, temperature 0.0), falling back to best-of-5 sampling across temperatures 0.2-1.0. Default settings were used for both WhisperX and Whisper from their respective repositories. Finally, to calculate WER, the \texttt{jiwer} Python package was used.\\

\subsection{Results and Discussion}
\begin{table*}[!ht] 
\centering
\caption{Evaluation of performance metrics (reported as percentages) for transcription variants. Comparison of WER, Pn Score, F1 and Speed on 15 transcripts from a long-form speech corpus. The arrows ($\uparrow$/$\downarrow$) indicate the preferred direction for each metric. Best performing values are highlighted in bold. }
\label{tab:earnings_results}
\begin{tabular*}{\textwidth}{@{\extracolsep{\fill}} l c r r r r r @{}}
\toprule
\textbf{Method} & \multicolumn{1}{c}{\textbf{\shortstack[b]{Condition \\ on Prev. Text}}} & \multicolumn{1}{c}{\textbf{WER $\downarrow$}} & \multicolumn{1}{c}{\textbf{Pn Score $\uparrow$}} & \multicolumn{1}{c}{\textbf{F1 $\uparrow$}} & \multicolumn{1}{c}{\textbf{LPS $\uparrow$}} & \multicolumn{1}{c}{\textbf{Speed $\uparrow$}} \\
\midrule
\textbf{WhisperX+IB (ours)} & \checkmark & \textbf{8.2} & \textbf{79.3} & 67.4 & \textbf{2.9} & 8.4$\times$ \\
WhisperX~\cite{bain2023} & $\times$ & 8.4 & 76.5 & 67.2 & 2.8 & \textbf{11.8}$\times$ \\
\midrule
\multirow{2}{*}{\texttt{openai/whisper}~\cite{radford2022}} & \checkmark & 12.0 & 72.0 & \textbf{67.9} & \textbf{2.9} & 1.0$\times$ \\
 & $\times$ & 11.9 & 66.9 & 66.7 & 2.8 & 1.0$\times$ \\
\bottomrule
\end{tabular*}
\end{table*}

In Table~\ref{tab:earnings_results}, we present the evaluation of our proposed Context-Aware Interleaved Batching (WhisperX+IB) against standard WhisperX and the sequential \texttt{openai/whisper} baseline on the Earnings-21 corpus. The results reveal that our method effectively bridges the performance gap caused by isolated batch processing. Although the sequential \texttt{openai/whisper} model with preceding text achieves a high F1 score (67.91) and a leading LPS of (2.9), it suffers from a high WER (12.0) and lacks parallelisation, operating at a 1.0× baseline speed. By re-enabling autoregressive context across chunk boundaries, WhisperX+IB matches the top LPS of (2.9), improves transcription accuracy and delivers an 8.4× speedup. Despite a slight drop in latency from the cold-start resolution pass, our method remains competitive with standard WhisperX (11.8×) while providing improved transcription structure and accuracy.\\

We can observe from the Earnings-21 results that \texttt{openai/whisper} outperforms our method in punctuation quality. WhisperX sets \texttt{without\_timestamps=True} for pure transcription (and therefore so does WhisperX+IB). The Whisper decoder is trained on two target formats controlled by a single conditioning token, one with interleaved timestamp tokens, and one without ~cite{radford2022}. We hypothesise that the timestamp-conditioned mode learns a tighter association between full stops and the tagged segment boundaries. Therefore, decoding without them may worsen the full stop placement and explain Whisper's higher F1 despite trailing on WER and Pn Score.

\begin{table}[!ht]
\centering
\small
\caption{Evaluation of transcription variants on medical YouTube lectures (reported as percentages). Comparison of WER and Pn Score, evaluated on a manually verified ground-truth dataset of medical terminology lesson transcripts. The arrows ($\uparrow$/$\downarrow$) indicate the preferred direction for each metric. Best performing values are highlighted in bold. Batch size 4 settings for WhisperX and WhisperX+IB.}
\label{tab:medical_results}
\begin{tabular*}{\columnwidth}{@{\extracolsep{\fill}} l r r @{}}
\toprule
\textbf{Variant} & \multicolumn{1}{c}{\textbf{WER $\downarrow$}} & \multicolumn{1}{c}{\textbf{Pn Score $\uparrow$}} \\
\midrule
\textbf{WhisperX+IB (ours)} & \textbf{3.3} & \textbf{84.7} \\
WhisperX                    & 3.5          & 83.6          \\
\bottomrule
\end{tabular*}
\end{table}

In Table~\ref{tab:medical_results}, we report results on the custom MedicalLessons dataset to evaluate the models' ability to handle complex, domain-specific terminology. We see that Interleaved Batching again outperforms the standard WhisperX approach. By successfully propagating preceding text context to subsequent batches, WhisperX+IB reduces the overall WER from 3.5\% to 3.3\%. More importantly, our proposed method yields a notable improvement in the Proper Noun (PN) Score, increasing it from 83.6\% to 84.7\%. This indicates that supplying contiguous textual context significantly aids the model in correctly predicting challenging medical vocabulary and technical terms that span across VAD-segmented chunk boundaries, a capability that is otherwise lost during isolated batched inference.\\ 

\begin{table}[!htbp]
\footnotesize
\caption{Qualitative comparison demonstrating the impact of context injection. Without access to historical context, the isolated \textit{WhisperX} model hallucinates the phrase ``post-2007''. In contrast, the \textit{WhisperX+IB} variant successfully leverages the preceding context to accurately transcribe ``upholstery fabric segment''.}
\label{tab:qualitative_example}
\hrule
\smallskip
\begin{flushleft}
\textbf{Preceding Context} \\
\textit{``For the \textbf{upholstery fabric segment}, sales for the third quarter were \$35 million down 5.7\% over the prior year, which prior year was a very strong quarter sales performance...''} \\[0.2cm]
\textbf{WhisperX} \\
\textit{``Return on capital for the trailing 12-month period for the \textbf{post-2007} continues to be impressive, coming in at 54\%. The home accessory segment, which includes our e-commerce...''} \\[0.2cm]
\textbf{WhisperX+IB (With preceding context)} \\
\textit{``Return on capital for the trailing 12-month period for the \textbf{upholstery fabric segment} continues to be impressive, coming in at 54\%. The home accessory segment, which includes our e-commerce...''}
\end{flushleft}
\hrule
\end{table}

Table~\ref{tab:qualitative_example}  illustrates the crucial role of context injection in improving transcription accuracy. When operating in isolation, the baseline WhisperX model hallucinates the phrase “post-2007.” By contrast, the WhisperX+IB variant successfully leverages preceding dialogue to contextually ground its output, eliminating the hallucination and accurately transcribing the phrase “upholstery fabric segment.”

\subsection{Limitations}
While Interleaved Batching is highly effective for long-form audio, an inherent limitation arises when the total number of VAD-segmented chunks ($N$) is less than or equal to the batch size ($B$). In such cases, transcription completes in a single parallel pass, precluding the propagation of preceding text context. Restoring this context requires artificially reducing the batch size, potentially down to $B=1$, which trades parallel throughput for contextual accuracy. Importantly, however, even in this fully sequential configuration ($B=1$), our approach maintains a significant accuracy advantage over standard Whisper. By decoupling autoregressive text conditioning from fragile timestamp decoding, our method avoids hallucination loops and preserves state-of-the-art WER and punctuation and proper noun accuracy.


\section{Conclusion}

In this paper, we address a major challenge in high-speed audio transcription: the trade-off between processing speed and accuracy. While modern systems achieve high throughput by breaking audio into separate chunks for parallel processing, this approach isolates the segments from one another, cutting off the continuous context and degrading transcription quality. Our proposed method successfully bridges this gap. The results demonstrate how WhisperX+IB reduces errors both grammatically and punctually. Ultimately, this framework proves that inference speed does not have to come at the expense of accuracy, offering a scalable solution for high-performance, context-aware speech transcription.

\clearpage
\bibliographystyle{IEEEtran}
\bibliography{references}

\end{document}